\documentclass[a4paper, 11pt]{article}
\usepackage[sort&compress,square,comma,authoryear]{natbib}

\usepackage{a4wide}
\usepackage{algorithm}
\usepackage{algorithmic}
\usepackage[algo2e,linesnumbered,ruled]{algorithm2e}

\usepackage{amsfonts}
\usepackage{amsmath}
\usepackage{amssymb}
\usepackage{amstext}
\usepackage{amsthm}
\usepackage{caption}
\usepackage{subcaption}
\usepackage{microtype}      
\usepackage{booktabs}
\usepackage{dirtytalk}
\usepackage{enumitem}
\usepackage{mathtools}
\usepackage{booktabs}
\usepackage[T1]{fontenc}
\usepackage{graphicx}
\usepackage{pifont}
\usepackage{setspace}
\usepackage{hyperref}
\usepackage[capitalize]{cleveref}

\usepackage{subcaption}
\usepackage{graphicx}

\usepackage{wrapfig}

\usepackage{thmtools}
\usepackage{thm-restate}

\usepackage{nicefrac}
\usepackage[dvipsnames]{xcolor} 
\usepackage[parfill]{parskip}

\definecolor{salmon}{RGB}{234,153,153}
\definecolor{cornflowerblue}{RGB}{6,69,173}
\hypersetup{
    colorlinks,
    linkcolor={cornflowerblue},
    citecolor={cornflowerblue},
    urlcolor={salmon}
}

\usepackage{tabularx}
\newcolumntype{Y}{>{\centering\arraybackslash}X}
\newcolumntype{Z}{>{\raggedleft\arraybackslash}X}
\usepackage{tikz}
\usetikzlibrary{calc, fadings, decorations, shapes, positioning, arrows}

\theoremstyle{plain}

\theoremstyle{definition}

\theoremstyle{remark}

\AfterEndEnvironment{theorem}{\noindent\ignorespaces}
\AfterEndEnvironment{proposition}{\noindent\ignorespaces}
\AfterEndEnvironment{lemma}{\noindent\ignorespaces}
\AfterEndEnvironment{corollary}{\noindent\ignorespaces}
\AfterEndEnvironment{definition}{\noindent\ignorespaces}
\AfterEndEnvironment{remark}{\noindent\ignorespaces}

\let\oldding\ding
\renewcommand{\ding}[2][1]{\scalebox{#1}{\oldding{#2}}}

\title{\textbf{Diffusion Based Causal Representation Learning}}

\date{}

\usepackage{authblk}

\author[1]{Amir Mohammad Karimi Mamaghan}
\author[2,3]{Andrea Dittadi}
\author[2,4]{\\Stefan Bauer}
\author[1]{Karl Henrik Johansson}
\author[5]{\\Francesco Quinzan}

\affil[1]{KTH Royal Institute of Technology}
\affil[2]{Helmholtz AI, Helmholtz Center Munich}
\affil[3]{Max Planck Institute for Intelligent Systems, Tübingen, Germany}
\affil[4]{Technical University of Munich}
\affil[5]{Department of Computer Science, University of Oxford}

\begin{document}

\maketitle

\begin{abstract}
  Causal reasoning can be considered a cornerstone of intelligent systems. Having access to an underlying causal graph comes with the promise of cause-effect estimation and the identification of efficient and safe interventions. However, 
  learning causal representations remains a major challenge, due to the complexity of many real-world systems. Previous works on causal representation learning have mostly focused on Variational Auto-Encoders (VAE). These methods only provide representations from a point estimate, and they are unsuitable to handle high dimensions. To overcome these problems, we proposed a new Diffusion-based Causal Representation Learning (DCRL) algorithm. This algorithm uses diffusion-based representations for causal discovery. DCRL offers access to infinite dimensional latent codes, which encode different levels of information in the latent code. In a first proof of principle, we investigate the use of DCRL for causal representation learning. We further demonstrate experimentally that this approach performs comparably well in identifying the causal structure and causal variables.
\end{abstract}

\section{Introduction}
Causal representation learning consists in uncovering a system's latent causal factors and their relationships, from observed low-level data. 
Causal representation learning finds applicability in domains such as autonomous driving \citep{scholkopf2021toward}, robotics \citep{hellstrom2021relevance}, healthcare~\citep{anwar2014multi}, climate studies~\citep{runge2019inferring}, epidemiology~\citep{hernan2000marginal, robins2000marginal}, and finance~\citep{hiemstra1994testing}. In these tasks, the underlying causal variables are often unknown, and we only have access to low-level representations. 

Causal representation learning is a challenging problem. In fact, identifying latent causal factors is generally impossible from observational data. There has been an ongoing effort to study sets of assumptions that ensure the identifiability of causal variables and their relationships \citep{yang2020causalvae, scholkopf2021toward, liu2022identifying, subramanian2022learning, brehmer2022weakly}. These approaches consider the availability of additional information or they use assumptions on the underlying causal structure of the DGP. Interestingly, \citet{brehmer2022weakly} consider a weak form of supervision in which we have access to a data pair, corresponding to the state of the system before and after a random, unknown intervention. \citet{brehmer2022weakly} prove that, in this weakly-supervised setting, the structure and the causal variables are identifiable up to a relabeling and element-wise reparameterization.

There has been a growing interest in leveraging generative models to learn causal representations with specific properties. For example, disentangled and object-centric representations have been shown to be helpful for complex downstream tasks and generalization \citep{van2019disentangled,dittadi2022generalization,wu2022slotformer,yoon2023investigation,papa2022inductive}. Variational Autoencoders (VAE) \citep{DBLP:journals/corr/KingmaW13} are among the most widely studied generative models, and they have been successfully used for disentanglement and causal representation learning \citep{locatello2020weakly,brehmer2022weakly}. However, the problem of learning causal representations has not yet been approached with more powerful generative models.

Recently, diffusion models have emerged as state-of-the-art generative models, and they have demonstrated remarkable success across several domains \citep{NEURIPS2021_49ad23d1, ramesh2022hierarchical, saharia2022photorealistic, ho2022video, höppe2022diffusion}. Diffusion models draw on concepts and principles from diffusion processes to learn the data distribution \citep{DBLP:journals/corr/MehrjouSS17,DBLP:journals/jmlr/HoSCFNS22,DBLP:journals/corr/abs-2101-02388,DBLP:conf/nips/DhariwalN21,DBLP:journals/corr/abs-1805-08306,DBLP:conf/iclr/ChenZZWNC21,DBLP:conf/eccv/CaiYAHBSH20,DBLP:conf/aistats/NiuSSZGE20,DBLP:conf/icml/SajjadiPMS18,DBLP:conf/iclr/0011SKKEP21,DBLP:conf/iclr/SongME21,DBLP:conf/icml/Sohl-DicksteinW15,sohl2015deep, ho2020denoising, song2021scorebased}. These models exploit diffusion behavior to produce diverse, high-quality, and realistic samples. Furthermore, diffusion-based models have the appealing property of infinite-dimensional latent codes \citep{abstreiter2022diffusionbased}, which allows to efficiently learn representations across different downstream tasks. Despite their remarkable performance and advantages, diffusion models have not yet been employed for causal representation learning, indicating that their potential has yet to be explored in this context.
\paragraph{Our contribution.} In this work, we study the connection between diffusion-based models and causal structure learning. In particular, our contributions are the following: 
\begin{itemize}
    \item We propose DCRL, a diffusion-based model for causal representation learning. We study and test the connection between the learned representations of DCRL with causal variables. To accomplish this, we utilize both finite and infinite-dimensional representations.
    \item We derive the Evidence Lower Bound (ELBO) for DCRL, in the case of both finite and infinite-dimensional representations. 
    \item We empirically illustrate that the noise and diffusion-based representations contain equivalent information about the underlying causal variables and causal mechanisms, and can be used interchangeably.
\end{itemize} 

\section{Related Work}
\paragraph{Diffusion-based Representation Learning.} Learning representations with diffusion models remains a relatively unexplored area. Several works try to train an external module (e.g., an encoder) along with the score function of the diffusion model to extract representations. \citet{abstreiter2022diffusionbased} and \citet{mittal2022points} condition the score function of a diffusion model on a time-independent and time-dependent encoder and obtain finite and infinite-dimensional representations, respectively. \citet{wang2023infodiffusion} use the same conditioning but regularizes the objective function with the mutual information between the input data and learned representations. \citet{traub2022representation} does the same conditioning but they use Latent Diffusion Models \citep{rombach2022high} where the inputs of the diffusion model are latent variables obtained from applying a pre-trained autoencoder on the input. Furthermore, \citet{kwon2022diffusion} proposes an asymmetric reverse process that discovers the semantic latent space of a frozen diffusion model where modification in the space synthesizes various attributes on input images. However, in principle, diffusion models lack a semantic latent space and it's unclear how to efficiently learn representations using their capabilities.

\paragraph{Causal Representation Learning.} Given the inherent challenges of identifiability in causal representation learning, many previous studies have tackled this issue by imposing certain assumptions on the dataset or the causal structure. Several previous methods rely on additional knowledge of the data generation process, such as knowledge of the causal graph or labels for the high-level causal variables. CausalGAN \citep{kocaoglu2017causalgan} requires the structure of the underlying causal graph to be known. \citet{yang2020causalvae} and \citet{liu2022identifying} assume a linear structural equation model, and they require additional information associated with the true causal concepts as supervising signals. Similar to \citet{yang2020causalvae}, \citet{komanduri2022scm} assume the availability of supplementary supervision labels, but without requiring mutual independence among factors. \citet{von2021self} investigate self-supervised causal representation learning by utilizing a known, but non-trivial, causal graph between content and style factors. \citet{subramanian2022learning} applies Bayesian structure learning in the latent space and relies on having interventional samples. 
For an overview of causal representation learning we refer to \citet{scholkopf2021toward}. 
Other relevant work closely related to causal representation learning includes disentangled representations and independent component analysis \citep{locatello2019challenging,shu2019weakly,lachapelle2022disentanglement,hyvarinen2000independent,khemakhem2020variational,ahuja2022weakly}.

\subsection{Overview}
The fundamental concept behind diffusion-based generative models is to learn to generate data by inverting a diffusion process. Diffusion models comprise two processes: a forward process and a backward process. The forward process gradually adds noise to data and maps data to (almost) pure noise.
The backward process, on the other hand, is used to go from a noise sample back to the original data space.

The forward process is defined by a stochastic differential equation (SDE) across a continuous time domain $t \in [0, 1]$, aiming to transform the data distribution to a known prior distribution, typically a standard multivariate Gaussian. Given $x_0$ sampled from a data distribution $p(x)$, the forward process constructs a trajectory $(x_t)_{t \in [0, 1]}$ across the time domain. We utilize the Variance Exploding SDE \citep{song2021scorebased} for the forward process, which is defined as:
\begin{equation*}
\label{eq:forward_sde}
    dx = f(x, t) + g(t)dw := \sqrt{\frac{d[\sigma^2(t)]}{dt}}dw,
\end{equation*}
where $w$ is the standard Wiener process and $\sigma^2(t)$ is the noise variance of the diffusion process at time $t$. The backward process is also formulated as an SDE in the following manner:
\begin{equation*}
    dx = [f(x,t) - g^2(t)\nabla_x\log p_t(x)]dt +g(t)d\Bar{w} \ ,
\end{equation*}
where $\Bar{w}$ is the standard Wiener process in reverse time. 
\paragraph{Score matching.} To use this backward process, the score function $\nabla_x \log p_t(x)$ is required. It is usually approximated by a neural score function $s_\theta(\cdot)$ which can be trained by Explicit Score Matching \citep{hyvarinen2005estimation} defined as:
\begin{align*}
   \mathcal{L}(\theta)  =  \mathbb{E}_t
   \Biggl[
       \lambda(t)
       \mathbb{E}_{p(x_t)}
       \Big[
          ||s_\theta(x_t, t)
            - \nabla_{x_t} \log p_t(x_t)||^2
       \Big]
   \Biggl],
\end{align*}
However, the ground-truth score function $\nabla_x \log p_t(x)$ is generally not known. \citet{vincent2011connection} addresses this issue by proposing Denoising Score Matching. The approximate score function is then learned by minimizing the loss function:
\begin{align*}
   \mathcal{L}(\theta)  = 
   \Biggl[        \lambda(t)
       \mathbb{E}_{x_0}\mathbb{E}_{p(x_t|x_0)}
       \Big[
           ||s_\theta(x_t, t)
            - \nabla_{x_t} \log p_t(x_t|x_0)||^2
       \Big]
   \Biggl],
\end{align*}
where the conditional distribution of $x_t$ given $x_0$ is 
$p_{t}(x_t|x_0) = \mathcal{N}(x_t;x_0, [\sigma^2(t) - \sigma^2(0)] \mathbf{I})$ and $\lambda(t)$ is a positive weighting function. This objective function originates from the evidence lower bound (ELBO) of the data distribution, and it's been shown that with a specific weighting function, this objective function becomes exactly a term in the ELBO \cite{song2021scorebased}. For more details, see Appendix \ref{appendix:elbo}.
\paragraph{Conditional Score Matching.} We can modify Denoising Score Matching to perform representation learning while training the score function. \citet{abstreiter2022diffusionbased} proposes conditional denoising score matching defined as:
\begin{align}
    \mathcal{L}(\theta, \phi)  = \mathbb{E}_t
    \Biggl[
        \lambda(t)
        \mathbb{E}_{x_0}\mathbb{E}_{p(x_t|x_0)} 
        \Big[
            ||s_\theta(x_t, E_\phi(x_0), t)
             - \nabla_{x_t} \log p_t(x_t|x_0)||^2
        \Big]
    \Biggl],
    \label{eq:diffusion_loss_single}
\end{align}
where the score function is conditioned on a module $E_\phi(x_0)$ which provides additional information about the data to the diffusion model through a learned encoder with parameters $\phi$. In fact, the encoder learns to extract necessary information from $x_0$ in a reduced-dimensional space that helps recover $x_0$ by denoising $x_t$.
\citet{abstreiter2022diffusionbased} also presents an alternative objective where the encoder is a function of time. Formally, the new objective is
\begin{align}
    \mathcal{L}(\theta, \phi)  = \mathbb{E}_t
    \Biggl[
        \lambda(t)
        \mathbb{E}_{x_0}\mathbb{E}_{p(x_t|x_0)} 
        \Big[
            ||s_\theta(x_t, E_\phi(x_0, t), t)
             - \nabla_{x_t} \log p_t(x_t|x_0)||^2
        \Big]
    \Biggl],
    \label{eq:diffusion_loss}
\end{align}
With this objective, the encoder learns a representation trajectory of $x_0$ instead of a single representation. Training this system has the potential to minimize the objective to zero, motivating the encoder $E_\phi(.)$ to learn meaningful, distinct representations at different timesteps \citep{abstreiter2022diffusionbased, mittal2022points}.
\subsection{Comparison with Other Generative Models}
The key difference between the other generative models and diffusion-based representations is that other generative models are only concerned with one finite code and all the information is encoded into this single code while in the latter, different levels of information are encoded along an infinite-dimensional code, i.e., the encoder is conditioned on time $t$ and produces a trajectory-based representation $(E_\phi(x_0, t))_{t \in [0, 1]}$. Within this representation, various points along the trajectory contain different levels of information, as highlighted by \citet{mittal2022points}.
In this work, we first explore a time-independent single code where we employ Eq. \ref{eq:diffusion_loss_single} and show that with a certain weighting function, this objective function will become the ELBO. Then, we apply the same experiments with infinite-dimensional latent code (Eq. \ref{eq:diffusion_loss}) and study the benefits and implications of these formulations for causal representation learning.
\section{Problem Description}

\begin{figure}[t]
    \centering
     \includegraphics[width=\textwidth]{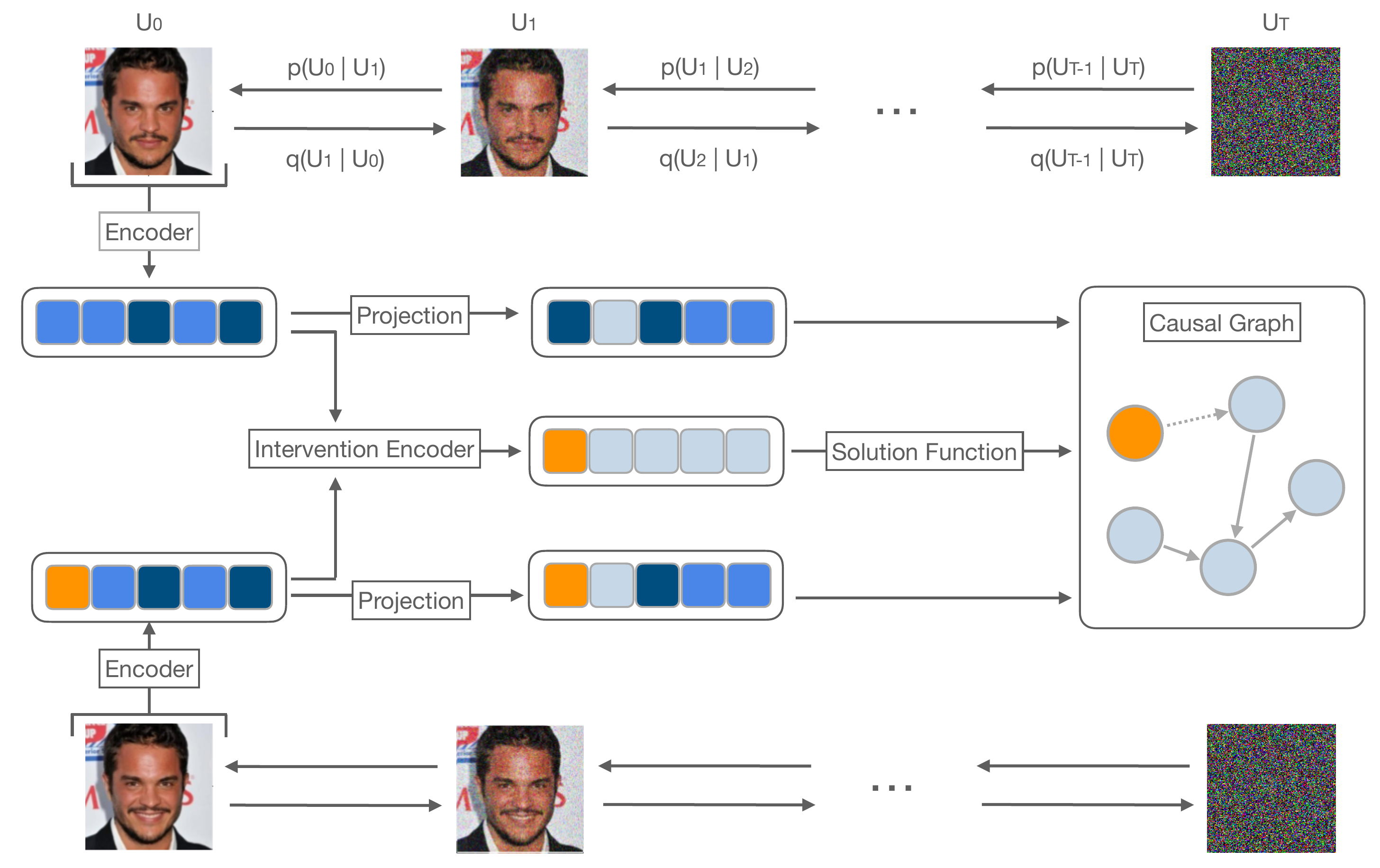}
     \caption{
     Overview of our framework. Here we have a paired image of a face before and after an intervention (the smile). The paired image is mapped to latent variables by a stochastic encoder. The intervention target is determined by applying the intervention encoder to these latent variables. To maintain the weakly supervised structure, the latent variables are projected into a new pair and then, serve as the conditioning module for a conditional diffusion model (The projected latent variables are diffusion-based representations of the input pair). Finally, they are utilized in neural solution functions together with the intervention target to obtain the latent causal variables.
     }
     \label{fig:framework}
\end{figure}

We consider a system that is described by an unknown underlying SCM on the latent causal variable $Z$ where we have access to low-level data pairs $(x, \Tilde{x}) \sim p(x, \Tilde{x})$ representing the system before and after a random, unknown, and atomic intervention. It is known that under this weakly supervised setting, it is possible to identify the causal variables and causal mechanisms up to a permutation and elementwise reparameterization of the variables \citep{brehmer2022weakly}. Our objective is to learn an SCM that accurately represents the true underlying SCM associated with the given data, up to a permutation and elementwise reparameterization of causal variables. To this end, we train an SCM by maximizing the likelihood of data. With sufficient data and perfect optimization, we can find the SCM that is equivalent to the ground-truth SCM.
\section{The DCRL Algorithm}

\subsection{Overview}

Figure \ref{fig:framework} provides a visual representation of the framework's architecture. In this study, we utilize a conditional diffusion model and apply it to the input data ($x$, $\Tilde{x}$) where $x, \Tilde{x} \in \mathbb{R}^{3 \times W \times H}$ and $W$ and $H$ are the width and height of the input, respectively. The conditioning module is defined as the encoding module, generating high-level diffusion-based representations $(e, \Tilde{e})$ for each low-level data pair where $e, \Tilde{e} \in \mathbb{R}^{d}$ and $d$ is the number of latent causal variables assumed to be known. We empirically show that these latent variables contain equivalent information as in noise variables of the underlying SCM and can be used interchangeably. Then, we infer the intervention target $I \in \{0, 1, ..., d-1\}$ for each data pair by an intervention module and use neural solution functions on top of the latent variables $(e, \Tilde{e})$ and the intervention target $I$ to obtain the underlying latent causal variable $(z, \Tilde{z})$. We describe each part in detail in the next paragraphs.
%
%
\paragraph{The Encoding and the Intervention Module.} The encoding module consists of two main parts: the \emph{stochastic encoder} and the \emph{projection module}. The stochastic encoder $q(e|x)$ maps data pairs ($x$, $\Tilde{x}$) to pre-projection latent variables ($e$, $\Tilde{e}$). The encoded inputs are then utilized in the intervention module $q(I|x, \Tilde{x})$ to infer the intervention target $I$ for the data pair ($x$, $\Tilde{x}$). Based on our data generation process, the encoded inputs have the property that only for the elements that are intervened upon, we have $e_i \neq \Tilde{e}_i, i \in I$, and the rest will remain the same. Based on this property, in order to infer interventions, we employ an intervention module $q(I|e, \Tilde{e})$ which is defined heuristically as
\begin{equation*}
    \log q(i \in I|x, \Tilde{x}) = \frac{1}{Z}(\alpha + \beta|\mu_e(x)_i - \mu_e(\Tilde{x})_i|
    + \gamma|\mu_e(x)_i - \mu_e(\Tilde{x})_i|^2)
\end{equation*}
Where $\mu_e(x)$ is the mean of the stochastic encoder $q(e|x)$, $\alpha$, $\beta$, and $\gamma$ are learnable parameters, and $Z$ is a normalization constant. Using this simple heuristic function, we increase the likelihood of a component as it undergoes more significant changes in response to interventions on the encoded input. Once the intervention is inferred from the pre-projection latent variables, we apply the projection module. The projection module is dependent on the inferred intervention target $I$ and projects the encoded input $(e, \Tilde{e})$ to new latent variables in a way that for the components $e_i$ that are not intervened upon, $i \notin I$, the pre-intervention and post-intervention latent components will be equal, $e_i = \Tilde{e}_i$. This prevents solution functions from deviating from the weakly supervised structure.

We write the combination of the encoder and the projection module as $q(e, \Tilde{e}|x, \Tilde{x}, I)$, and refer it to as the \emph{encoding module}. By this definition, the encoding module $q(e, \Tilde{e}|x, \Tilde{x}, I)$ maps the input ($x$, $\Tilde{x}$) to latent variables ($e$, $\Tilde{e}$) and the intervention module infers the intervention $I$ based on pre-projection latent variables.
\paragraph{Prior.} Given the intervention target $I$ and latent variables ($e$, $\Tilde{e}$), we define the prior $p(e, \Tilde{e}, I)$ as $p(e, \Tilde{e}, I) = p(I)p(e)p(\Tilde{e}|e, I)$. The objective of the prior distribution is to implicitly capture the causal structure and causal mechanisms within the system. Specifically, $p(I)$ and $p(e)$ denote the prior distributions over intervention targets and latent variables, respectively, and are configured as uniform categorical and standard Gaussian distributions, respectively. According to our data generation process, when an intervention is applied, only the elements in the latent variables that are intervened upon are altered; the other elements remain unchanged and independent of each other. Consequently, we can define $p(\Tilde{e}|e, I)$ as follows:
\begin{equation*}
    \label{eq:ilcm_prior}
    p(\Tilde{e}|e, I) = \prod_{i \notin I}\delta(\Tilde{e}_i - e_i) 
    \prod_{i \in I}p(\Tilde{e}_i|e)
\end{equation*}
In this equation, $\delta(.)$ is the Dirac delta function that fulfills this property for non-intervened elements of latent variables. 
\paragraph{Neural Solution Functions.} Finally, in order to encode the information about the intervened variables, we incorporate a conditional normalizing flow $p(\Tilde{e}_i|e)$ defined as
\begin{equation*}
    p(\Tilde{e}_i|e) = \Tilde{p}(h_i(\Tilde{e}_i; e_{\\i}))\Big|
        \frac{\partial h_i(\Tilde{e}_i; e_{i})}{\partial \Tilde{e}_i}
    \Big|
\end{equation*}
where $h(.)$ are the solution functions of the SCM. They are defined as invertible affine transformations with parameters learned with neural networks. Therefore, by learning solution functions, i.e., learning to transform $e$ to $z$, we implicitly model the causal graph into the framework and obtain the latent causal variables. For more details about the implementation, see Appendix \ref{appendix:implementation_details}.
\subsection{The Evidence Lower Bound for DCRL}
Putting everything together, we calculate the Evidence Lower Bound (ELBO) for the proposed model which will be:
\begin{align*}
    \log p(x, \Tilde{x}) & \geq
    \mathbb{E}_{p(x,\Tilde{x})} 
        \mathbb{E}_{q(I|x,\Tilde{x})} \mathbb{E}_{q(e, \Tilde{e}|x, \Tilde{x}, I)} \mathbb{E}_{t \sim U(0, 1)}
            \mathbb{E}_{q(u_t|x)}\mathbb{E}_{q(\Tilde{u}_t|\Tilde{x})}
            \Biggl[
                \log p(I) + \log p(e) \\
                +& \log p(\Tilde{e}|e, I)
                - \log q(I|x, \Tilde{x}) 
                - \log q(e, \Tilde{e}|x, \Tilde{x}, I)
                +
                \lambda(t)||s_\theta(u_t, e, t) - \nabla_{u_t} \log p(u_t|x)||_2^2 \\
                +&
                \lambda(t)||s_\theta(\Tilde{u}_t, \Tilde{e}, t) - \nabla_{\Tilde{u}_t} \log p(\Tilde{u}_t|\Tilde{x})||_2^2
            \Biggl],
\end{align*}
where $\lambda(t)$ is a positive weighting function. We train the model by minimizing a reweighted loss function reminiscent of $\beta$-VAEs:
\begin{align*}
    \mathcal{L}_{model}& =
    \mathbb{E}_{p(x,\Tilde{x})} 
        \mathbb{E}_{q(I|x,\Tilde{x})} \mathbb{E}_{q(e, \Tilde{e}|x, \Tilde{x}, I)}
            \mathbb{E}_{t \sim U(0, 1)} \mathbb{E}_{q(u_t|x)}\mathbb{E}_{q(\Tilde{u}_t|\Tilde{x})}
            \Biggl[
                \lambda(t)||s_\theta(u_t, e, t) \\ &- \nabla_{u_t} \log p(u_t|x)||_2^2
                +
                \lambda(t)||s_\theta(\Tilde{u}_t, \Tilde{e}, t) - \nabla_{\Tilde{u}_t} \log p(\Tilde{u}_t|\Tilde{x})||_2^2
                +
                \beta \Big[
                    \log p(I) + \log p(e) \\ 
                    &+ \log p(\Tilde{e}|e, I)
                    -
                    \log q(I|x, \Tilde{x}) - \log q(e, \Tilde{e}|x, \Tilde{x}, I) 
                \Big]
            \Biggl],
\end{align*}
In case of using infinite-dimensional representations (Eq. \ref{eq:diffusion_loss}), the objective function becomes:
\begin{align}
    \mathcal{L}_{model}& =
    \mathbb{E}_{p(x,\Tilde{x})} 
        \mathbb{E}_{q(I|x,\Tilde{x})} \mathbb{E}_{t \sim U(0, 1)} \mathbb{E}_{q(e_t, \Tilde{e_t}|x, \Tilde{x}, I)}
            \mathbb{E}_{q(u_t|x)}\mathbb{E}_{q(\Tilde{u}_t|\Tilde{x})}
            \Biggl[
                \lambda(t)||s_\theta(u_t, e_t, t) \nonumber
                \\ &- \nabla_{u_t} \log p(u_t|x)||_2^2
                +
                \lambda(t)||s_\theta(\Tilde{u}_t, \Tilde{e_t}, t) - \nabla_{\Tilde{u}_t} \log p(\Tilde{u}_t|\Tilde{x})||_2^2
                +
                \beta \Big[
                    \log p(I) + \log p(e_t) \nonumber
                    \\ 
                    &+ \log p(\Tilde{e}_t|e_t, I)
                    -
                    \log q(I|x, \Tilde{x}) - \log q(e_t, \Tilde{e}_t|x, \Tilde{x}, I) 
                \Big]
            \Biggl],
            \label{eq:loss}
\end{align}
where $(e_t)_{t \in [0, 1]}$ is the trajectory-based representation and $e_t \in \mathbb{R}^d$ is the single point of the trajectory at time $t$. For more details about the problem formulation, see Appendix \ref{appendix:elbo}. To prevent a collapse of the latent space to a lower-dimensional subspace, we add the negative entropy of the batch-aggregate intervention posterior ($q_I^{batch}(I) = \mathbb{E}_{x, \Tilde{x} \in batch}[q(I|x, \Tilde{x}]$) as a regularization term to the loss function:
\begin{equation*}
    \mathcal{L}_{entropy} = \mathbb{E}_{batches}
    \Big[
        - \sum_I q_I^{batch}(I) \log q_I^{batch}(I)
    \Big]
\end{equation*}
where $\mathbb{E}_{batches}[\ \cdot \ ]$ is the expected value over all the batches of data. After the training, the framework contains information about the underlying causal structure and latent causal variables and it can be used in different downstream tasks.
\section{Experiments}
\subsection{Overview of the Experiments}
Here we analyze the performance of the proposed model, DCRL, on synthetic data. We employ DCRL for the task of causal discovery and subsequently use ENCO \citep{lippe2021efficient}, a continuous optimization structure learning method that leverages observational and interventional data, on top of DCRL to infer the underlying causal graph. Furthermore, we evaluate the learned latent variables with the DCI framework \citep{eastwood2018framework}.
\begin{figure}[t]
    \centering
    \includegraphics[width=0.95\textwidth]{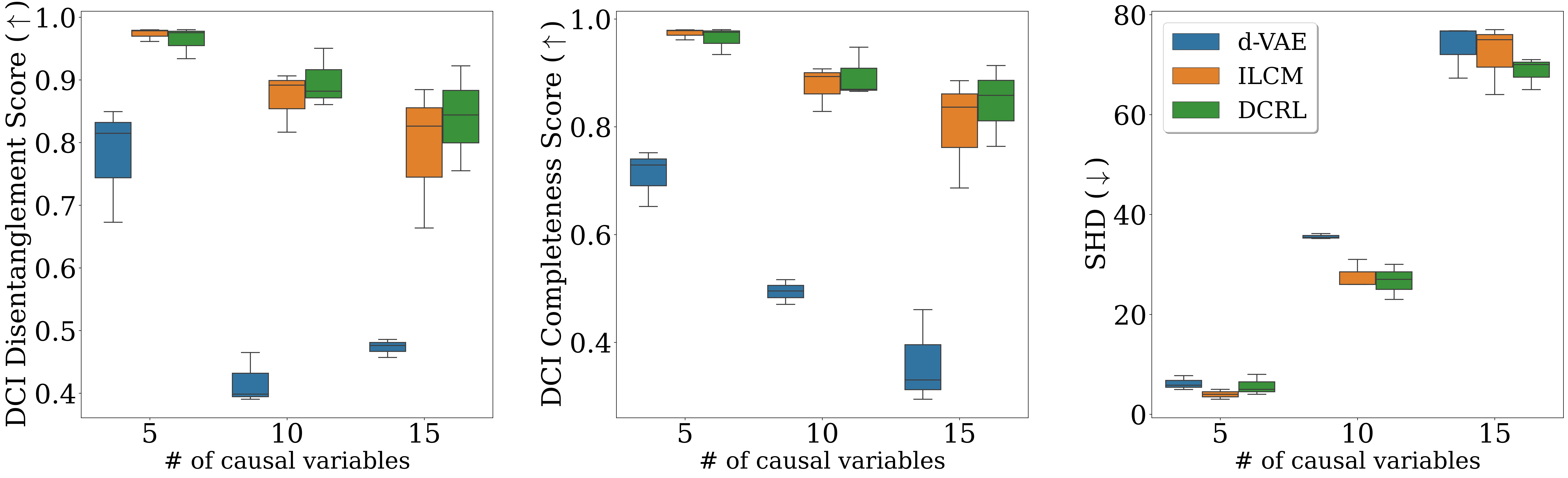}
    \caption{Comparison of models on different metrics when using single-point representation. Our approach outperforms or competes favorably with the baseline methods on all metrics. Particularly in higher dimensions, our method excels by capturing additional information about the causal variables and the underlying causal structure.}
    \label{fig:metrics}
\end{figure}
\paragraph{Data Generation.} In order to generate latent variables, we adopt random graphs where each edge in a fixed topological order is sampled from a Bernoulli distribution with a parameter that is equal to 0.5. We consider the SCM to be linear Gaussian and we sample the weights from a multivariate Normal distribution with zero mean and unit variance. We make sure the weights are not close to zero to avoid the violation of the faithfulness assumption. We introduce additive Gaussian noise with equal variances across all nodes, with its variance set to 0.1. Latent causal variables are then sampled using ancestral sampling, and we generate $10^5$ training samples, $10^4$ validation samples, and $10^4$ test samples. Finally, to generate input data $x$, we apply a random linear projection on the obtained latent variables. We keep the dimension of $x$ fixed to 16. We utilize an SCM with 5, 10, and 15 variables. To enhance the robustness of the results, we generate data for 4 different seeds and repeat our experiments for each seed.\looseness=-1
\paragraph{Baselines.} We consider ILCM as our main baseline. To the best of our knowledge, there aren't any other methods that consider the same weakly-supervised assumptions. We also evaluate the outcomes against a variation of disentanglement VAE proposed by \citep{locatello2020weakly} tailored for weakly supervised settings. This model, referred to as d-VAE, models the weakly supervised process but assumes unconnected variation factors instead of a causal relationship among variables. Similarly, we apply ENCO on top of both to obtain the learned graph.
\paragraph{Metrics.} We assess the performance of models with the following metrics:
\begin{itemize}[itemsep=4pt,parsep=0pt]
    \item \textbf{Structural Hamming Distance (SHD)}~ is a metric used to quantify the dissimilarity between two directed acyclic graphs (DAGs) by measuring the minimum number of edge additions, deletions, and reversals required to transform one graph into another. It is calculated by summing up the absolute differences between the entries of adjacency matrices of two graphs.
    \item \textbf{DCI Disentanglement Score}~ is a metric used to evaluate the disentanglement quality of a generative model and takes values between $0$ and $1$. Disentanglement refers to the extent to which the model learns to predict the underlying factors of variation in the data in a way that each predicted variable captures at most one underlying factor. If a predicted factor is important to predict a single underlying factor, the score will be $1$, and if a predicted factor is equally important to predict all the underlying factors, the score will be $0$ \citep{eastwood2018framework}.
    \item \textbf{DCI Completeness Score}~ measures how well each underlying factor of variation is captured by a single predicted latent variable and has a value between $0$ and $1$. If a single variable contributes to one underlying factor, the score will be 1, and if all variables equally contribute to the prediction of a single factor, the score will be 0 \citep{eastwood2018framework}.
\end{itemize}
\subsection{Single-point Representations}
Utilizing single-point representations where $e \in \mathbb{R}^d$ and is independent of time, our method demonstrates superior or competitive performance compared to the baselines, as indicated by the metrics shown in Figure \ref{fig:metrics}. In higher dimensions, our method excels by acquiring more information about the causal variables and underlying causal structure.
\subsection{Infinite-dimensional Representations}
In these experiments, we utilize the infinite-dimensional representations approach and develop trajectory-based representations for each input $x_0$, denoted as $(e_t)_{t \in [0, 1]}$. In order to perform inference, we sample points from this trajectory at intervals of $0.1$ resulting in $11$ specific time steps. The outcomes are depicted in Figure \ref{fig:metrics_infinite_dimensional} in Appendix \ref{appendix:missing_figures}. Generally, representations in the middle of the trajectory contain the most information and are comparable to or even outperform the baselines. Going further in time, representations appear to lose information but improve as they move towards the end of the trajectory. This phenomenon arises because during training, as we are further in time, the noise in the diffusion model is pretty high and the conditioning module compensates for that by providing the necessary information for the diffusion model to learn the score function.
\section{Conclusion}
Identifying the underlying causal variables and mechanisms of a system solely from observational data is considered impossible without additional assumptions. In this project, we use weak supervision as an inductive bias and study if the information encoded in the latent code of diffusion-based representations contains useful knowledge of causal variables and the underlying causal graph.

\bibliographystyle{abbrvnat}
\bibliography{bibliography.bib}

\begin{thebibliography}{58}
\providecommand{\natexlab}[1]{#1}
\providecommand{\url}[1]{\texttt{#1}}
\expandafter\ifx\csname urlstyle\endcsname\relax
  \providecommand{\doi}[1]{doi: #1}\else
  \providecommand{\doi}{doi: \begingroup \urlstyle{rm}\Url}\fi

\bibitem[Abstreiter et~al.(2022)Abstreiter, Mittal, Bauer, Schölkopf, and
  Mehrjou]{abstreiter2022diffusionbased}
K.~Abstreiter, S.~Mittal, S.~Bauer, B.~Schölkopf, and A.~Mehrjou.
\newblock Diffusion-based representation learning.
\newblock \emph{CoRR}, abs/2105.14257, 2022.

\bibitem[Ahuja et~al.(2022)Ahuja, Hartford, and Bengio]{ahuja2022weakly}
K.~Ahuja, J.~S. Hartford, and Y.~Bengio.
\newblock Weakly supervised representation learning with sparse perturbations.
\newblock \emph{Proc. of {N}eur{IPS}}, 35:\penalty0 15516--15528, 2022.

\bibitem[Anwar et~al.(2014)Anwar, Mideska, Hellriegel, Hoogenboom, Krause,
  Schnitzler, Deuschl, Raethjen, Heute, and Muthuraman]{anwar2014multi}
A.~R. Anwar, K.~G. Mideska, H.~Hellriegel, N.~Hoogenboom, H.~Krause,
  A.~Schnitzler, G.~Deuschl, J.~Raethjen, U.~Heute, and M.~Muthuraman.
\newblock Multi-modal causality analysis of eyes-open and eyes-closed data from
  simultaneously recorded eeg and meg.
\newblock In \emph{Proc. of {EMBC}}, pages 2825--2828, 2014.

\bibitem[Brehmer et~al.(2022)Brehmer, De~Haan, Lippe, and
  Cohen]{brehmer2022weakly}
J.~Brehmer, P.~De~Haan, P.~Lippe, and T.~S. Cohen.
\newblock Weakly supervised causal representation learning.
\newblock In \emph{Proc. of {N}eur{IPS}}, pages 2256--2265, 2022.

\bibitem[Cai et~al.(2020)Cai, Yang, Averbuch{-}Elor, Hao, Belongie, Snavely,
  and Hariharan]{DBLP:conf/eccv/CaiYAHBSH20}
R.~Cai, G.~Yang, H.~Averbuch{-}Elor, Z.~Hao, S.~J. Belongie, N.~Snavely, and
  B.~Hariharan.
\newblock Learning gradient fields for shape generation.
\newblock In \emph{Proc. fo {ECCV}}, volume 12348, pages 364--381, 2020.

\bibitem[Chen et~al.(2021)Chen, Zhang, Zen, Weiss, Norouzi, and
  Chan]{DBLP:conf/iclr/ChenZZWNC21}
N.~Chen, Y.~Zhang, H.~Zen, R.~J. Weiss, M.~Norouzi, and W.~Chan.
\newblock Wavegrad: Estimating gradients for waveform generation.
\newblock In \emph{Proc. of {ICLR}}, 2021.

\bibitem[Dhariwal and Nichol(2021{\natexlab{a}})]{NEURIPS2021_49ad23d1}
P.~Dhariwal and A.~Nichol.
\newblock Diffusion models beat gans on image synthesis.
\newblock In \emph{Proc. of {N}eur{IPS}}, pages 8780--8794, 2021{\natexlab{a}}.

\bibitem[Dhariwal and Nichol(2021{\natexlab{b}})]{DBLP:conf/nips/DhariwalN21}
P.~Dhariwal and A.~Q. Nichol.
\newblock Diffusion models beat gans on image synthesis.
\newblock In \emph{Proc. of {N}eur{IPS}}, pages 8780--8794, 2021{\natexlab{b}}.

\bibitem[Dittadi et~al.(2022)Dittadi, Papa, De~Vita, Sch{\"o}lkopf, Winther,
  and Locatello]{dittadi2022generalization}
A.~Dittadi, S.~Papa, M.~De~Vita, B.~Sch{\"o}lkopf, O.~Winther, and
  F.~Locatello.
\newblock Generalization and robustness implications in object-centric
  learning.
\newblock In \emph{Proc. of {ICML}}, pages 5221--5285, 2022.

\bibitem[Eastwood and Williams(2018)]{eastwood2018framework}
C.~Eastwood and C.~K. Williams.
\newblock A framework for the quantitative evaluation of disentangled
  representations.
\newblock In \emph{Proc. of {ICLR}}, 2018.

\bibitem[Hellstr{\"o}m(2021)]{hellstrom2021relevance}
T.~Hellstr{\"o}m.
\newblock The relevance of causation in robotics: A review, categorization, and
  analysis.
\newblock \emph{Paladyn, Journal of Behavioral Robotics}, 12\penalty0
  (1):\penalty0 238--255, 2021.

\bibitem[Hern{\'a}n et~al.(2000)Hern{\'a}n, Brumback, and
  Robins]{hernan2000marginal}
M.~{\'A}. Hern{\'a}n, B.~Brumback, and J.~M. Robins.
\newblock Marginal structural models to estimate the causal effect of
  zidovudine on the survival of hiv-positive men.
\newblock \emph{Epidemiology}, pages 561--570, 2000.

\bibitem[Hiemstra and Jones(1994)]{hiemstra1994testing}
C.~Hiemstra and J.~D. Jones.
\newblock Testing for linear and nonlinear granger causality in the stock
  price-volume relation.
\newblock \emph{The Journal of Finance}, 49\penalty0 (5):\penalty0 1639--1664,
  1994.

\bibitem[Ho et~al.(2020)Ho, Jain, and Abbeel]{ho2020denoising}
J.~Ho, A.~Jain, and P.~Abbeel.
\newblock Denoising diffusion probabilistic models.
\newblock In \emph{Proc. of {N}eur{IPS}}, pages 8780--8794, 2020.

\bibitem[Ho et~al.(2022{\natexlab{a}})Ho, Saharia, Chan, Fleet, Norouzi, and
  Salimans]{DBLP:journals/jmlr/HoSCFNS22}
J.~Ho, C.~Saharia, W.~Chan, D.~J. Fleet, M.~Norouzi, and T.~Salimans.
\newblock Cascaded diffusion models for high fidelity image generation.
\newblock \emph{Journal of Machine Learning Research}, 23:\penalty0
  47:1--47:33, 2022{\natexlab{a}}.

\bibitem[Ho et~al.(2022{\natexlab{b}})Ho, Salimans, Gritsenko, Chan, Norouzi,
  and Fleet]{ho2022video}
J.~Ho, T.~Salimans, A.~Gritsenko, W.~Chan, M.~Norouzi, and D.~J. Fleet.
\newblock Video diffusion models.
\newblock \emph{CoRR}, abs/2204.03458, 2022{\natexlab{b}}.

\bibitem[Hyv{\"a}rinen and Dayan(2005)]{hyvarinen2005estimation}
A.~Hyv{\"a}rinen and P.~Dayan.
\newblock Estimation of non-normalized statistical models by score matching.
\newblock \emph{Journal of Machine Learning Research}, 6\penalty0 (4), 2005.

\bibitem[Hyv{\"a}rinen and Oja(2000)]{hyvarinen2000independent}
A.~Hyv{\"a}rinen and E.~Oja.
\newblock Independent component analysis: algorithms and applications.
\newblock \emph{Neural networks}, 13\penalty0 (4-5):\penalty0 411--430, 2000.

\bibitem[Höppe et~al.(2022)Höppe, Mehrjou, Bauer, Nielsen, and
  Dittadi]{höppe2022diffusion}
T.~Höppe, A.~Mehrjou, S.~Bauer, D.~Nielsen, and A.~Dittadi.
\newblock Diffusion models for video prediction and infilling.
\newblock \emph{CoRR}, abs/2206.07696, 2022.

\bibitem[Khemakhem et~al.(2020)Khemakhem, Kingma, Monti, and
  Hyvarinen]{khemakhem2020variational}
I.~Khemakhem, D.~Kingma, R.~Monti, and A.~Hyvarinen.
\newblock Variational autoencoders and nonlinear ica: A unifying framework.
\newblock In \emph{Proc. of {AISTATS}}, pages 2207--2217, 2020.

\bibitem[Kingma and Welling(2014)]{DBLP:journals/corr/KingmaW13}
D.~P. Kingma and M.~Welling.
\newblock Auto-encoding variational bayes.
\newblock In Y.~Bengio and Y.~LeCun, editors, \emph{Proc. of {ICLR}}, 2014.

\bibitem[Kocaoglu et~al.(2017)Kocaoglu, Snyder, Dimakis, and
  Vishwanath]{kocaoglu2017causalgan}
M.~Kocaoglu, C.~Snyder, A.~G. Dimakis, and S.~Vishwanath.
\newblock ausalgan: Learning causal implicit generative models with adversarial
  training.
\newblock \emph{CoRR}, abs/1709.02023, 2017.

\bibitem[Komanduri et~al.(2022)Komanduri, Wu, Huang, Chen, and
  Wu]{komanduri2022scm}
A.~Komanduri, Y.~Wu, W.~Huang, F.~Chen, and X.~Wu.
\newblock Scm-vae: Learning identifiable causal representations via structural
  knowledge.
\newblock In \emph{{IEEE} {B}ig {D}ata}, pages 1014--1023, 2022.

\bibitem[Kwon et~al.(2022)Kwon, Jeong, and Uh]{kwon2022diffusion}
M.~Kwon, J.~Jeong, and Y.~Uh.
\newblock Diffusion models already have a semantic latent space.
\newblock \emph{CoRR}, abs/2210.10960, 2022.

\bibitem[Lachapelle et~al.(2022)Lachapelle, Rodriguez, Sharma, Everett,
  Le~Priol, Lacoste, and Lacoste-Julien]{lachapelle2022disentanglement}
S.~Lachapelle, P.~Rodriguez, Y.~Sharma, K.~E. Everett, R.~Le~Priol, A.~Lacoste,
  and S.~Lacoste-Julien.
\newblock Disentanglement via mechanism sparsity regularization: A new
  principle for nonlinear ica.
\newblock In \emph{Proc. of {CLR}}, pages 428--484, 2022.

\bibitem[Lippe et~al.(2021)Lippe, Cohen, and Gavves]{lippe2021efficient}
P.~Lippe, T.~Cohen, and E.~Gavves.
\newblock Efficient neural causal discovery without acyclicity constraints.
\newblock \emph{CoRR}, abs/2107.10483, 2021.

\bibitem[Liu et~al.(2022)Liu, Zhang, Gong, Gong, Huang, Hengel, Zhang, and
  Shi]{liu2022identifying}
Y.~Liu, Z.~Zhang, D.~Gong, M.~Gong, B.~Huang, A.~v.~d. Hengel, K.~Zhang, and
  J.~Q. Shi.
\newblock Identifying weight-variant latent causal models.
\newblock \emph{CoRR}, abs/2208.14153, 2022.

\bibitem[Locatello et~al.(2019)Locatello, Bauer, Lucic, Raetsch, Gelly,
  Sch{\"o}lkopf, and Bachem]{locatello2019challenging}
F.~Locatello, S.~Bauer, M.~Lucic, G.~Raetsch, S.~Gelly, B.~Sch{\"o}lkopf, and
  O.~Bachem.
\newblock Challenging common assumptions in the unsupervised learning of
  disentangled representations.
\newblock In \emph{Proc. of {ICML}}, pages 4114--4124, 2019.

\bibitem[Locatello et~al.(2020)Locatello, Poole, R{\"a}tsch, Sch{\"o}lkopf,
  Bachem, and Tschannen]{locatello2020weakly}
F.~Locatello, B.~Poole, G.~R{\"a}tsch, B.~Sch{\"o}lkopf, O.~Bachem, and
  M.~Tschannen.
\newblock Weakly-supervised disentanglement without compromises.
\newblock In \emph{Proc. of {ICML}}, pages 6348--6359, 2020.

\bibitem[Luhman and Luhman(2021)]{DBLP:journals/corr/abs-2101-02388}
E.~Luhman and T.~Luhman.
\newblock Knowledge distillation in iterative generative models for improved
  sampling speed.
\newblock \emph{CoRR}, abs/2101.02388, 2021.

\bibitem[Luo(2022)]{luo2022understanding}
C.~Luo.
\newblock Understanding diffusion models: A unified perspective.
\newblock \emph{CoRR}, abs/2208.11970, 2022.

\bibitem[Mehrjou et~al.(2017)Mehrjou, Sch{\"{o}}lkopf, and
  Saremi]{DBLP:journals/corr/MehrjouSS17}
A.~Mehrjou, B.~Sch{\"{o}}lkopf, and S.~Saremi.
\newblock Annealed generative adversarial networks.
\newblock \emph{CoRR}, abs/1705.07505, 2017.

\bibitem[Mittal et~al.(2022)Mittal, Lajoie, Bauer, and
  Mehrjou]{mittal2022points}
S.~Mittal, G.~Lajoie, S.~Bauer, and A.~Mehrjou.
\newblock From points to functions: Infinite-dimensional representations in
  diffusion models, 2022.

\bibitem[Niu et~al.(2020)Niu, Song, Song, Zhao, Grover, and
  Ermon]{DBLP:conf/aistats/NiuSSZGE20}
C.~Niu, Y.~Song, J.~Song, S.~Zhao, A.~Grover, and S.~Ermon.
\newblock Permutation invariant graph generation via score-based generative
  modeling.
\newblock In \emph{Proc. of {AISTATS}}, volume 108, pages 4474--4484, 2020.

\bibitem[Papa et~al.(2022)Papa, Winther, and Dittadi]{papa2022inductive}
S.~Papa, O.~Winther, and A.~Dittadi.
\newblock Inductive biases for object-centric representations in the presence
  of complex textures.
\newblock In \emph{UAI 2022 Workshop on Causal Representation Learning}, 2022.

\bibitem[Ramesh et~al.(2022)Ramesh, Dhariwal, Nichol, Chu, and
  Chen]{ramesh2022hierarchical}
A.~Ramesh, P.~Dhariwal, A.~Nichol, C.~Chu, and M.~Chen.
\newblock Hierarchical text-conditional image generation with clip latents.
\newblock \emph{CoRR}, abs/2204.06125, 2022.

\bibitem[Robins et~al.(2000)Robins, Hernan, and Brumback]{robins2000marginal}
J.~M. Robins, M.~A. Hernan, and B.~Brumback.
\newblock Marginal structural models and causal inference in epidemiology.
\newblock \emph{Epidemiology}, pages 550--560, 2000.

\bibitem[Rombach et~al.(2022)Rombach, Blattmann, Lorenz, Esser, and
  Ommer]{rombach2022high}
R.~Rombach, A.~Blattmann, D.~Lorenz, P.~Esser, and B.~Ommer.
\newblock High-resolution image synthesis with latent diffusion models.
\newblock In \emph{Proc. of {ECCV}}, pages 10684--10695, 2022.

\bibitem[Runge et~al.(2019)Runge, Bathiany, Bollt, Camps-Valls, Coumou, Deyle,
  Glymour, Kretschmer, Mahecha, Mu{\~n}oz-Mar{\'\i},
  et~al.]{runge2019inferring}
J.~Runge, S.~Bathiany, E.~Bollt, G.~Camps-Valls, D.~Coumou, E.~Deyle,
  C.~Glymour, M.~Kretschmer, M.~D. Mahecha, J.~Mu{\~n}oz-Mar{\'\i}, et~al.
\newblock Inferring causation from time series in earth system sciences.
\newblock \emph{Nature Communications}, 10\penalty0 (1):\penalty0 2553, 2019.

\bibitem[Saharia et~al.(2022)Saharia, Chan, Saxena, Li, Whang, Denton,
  Ghasemipour, Gontijo~Lopes, Karagol~Ayan, Salimans,
  et~al.]{saharia2022photorealistic}
C.~Saharia, W.~Chan, S.~Saxena, L.~Li, J.~Whang, E.~L. Denton, K.~Ghasemipour,
  R.~Gontijo~Lopes, B.~Karagol~Ayan, T.~Salimans, et~al.
\newblock Photorealistic text-to-image diffusion models with deep language
  understanding.
\newblock In \emph{Proc. of {N}eur{IPS}}, 2022.

\bibitem[Sajjadi et~al.(2018)Sajjadi, Parascandolo, Mehrjou, and
  Sch{\"{o}}lkopf]{DBLP:conf/icml/SajjadiPMS18}
M.~S.~M. Sajjadi, G.~Parascandolo, A.~Mehrjou, and B.~Sch{\"{o}}lkopf.
\newblock Tempered adversarial networks.
\newblock In \emph{Proc. of {ICML}}, volume~80, pages 4448--4456, 2018.

\bibitem[Saremi et~al.(2018)Saremi, Mehrjou, Sch{\"{o}}lkopf, and
  Hyv{\"{a}}rinen]{DBLP:journals/corr/abs-1805-08306}
S.~Saremi, A.~Mehrjou, B.~Sch{\"{o}}lkopf, and A.~Hyv{\"{a}}rinen.
\newblock Deep energy estimator networks.
\newblock \emph{CoRR}, abs/1805.08306, 2018.

\bibitem[Sch{\"o}lkopf et~al.(2021)Sch{\"o}lkopf, Locatello, Bauer, Ke,
  Kalchbrenner, Goyal, and Bengio]{scholkopf2021toward}
B.~Sch{\"o}lkopf, F.~Locatello, S.~Bauer, N.~R. Ke, N.~Kalchbrenner, A.~Goyal,
  and Y.~Bengio.
\newblock Toward causal representation learning.
\newblock \emph{Proceedings of the {IEEE}}, 109\penalty0 (5):\penalty0
  612--634, 2021.

\bibitem[Shu et~al.(2019)Shu, Chen, Kumar, Ermon, and Poole]{shu2019weakly}
R.~Shu, Y.~Chen, A.~Kumar, S.~Ermon, and B.~Poole.
\newblock Weakly supervised disentanglement with guarantees.
\newblock \emph{arXiv preprint arXiv:1910.09772}, 2019.

\bibitem[Sohl-Dickstein et~al.(2015)Sohl-Dickstein, Weiss, Maheswaranathan, and
  Ganguli]{sohl2015deep}
J.~Sohl-Dickstein, E.~Weiss, N.~Maheswaranathan, and S.~Ganguli.
\newblock Deep unsupervised learning using nonequilibrium thermodynamics.
\newblock In \emph{Proc. of {ICML}}, pages 2256--2265, 2015.

\bibitem[Sohl{-}Dickstein et~al.(2015)Sohl{-}Dickstein, Weiss, Maheswaranathan,
  and Ganguli]{DBLP:conf/icml/Sohl-DicksteinW15}
J.~Sohl{-}Dickstein, E.~A. Weiss, N.~Maheswaranathan, and S.~Ganguli.
\newblock Deep unsupervised learning using nonequilibrium thermodynamics.
\newblock In \emph{Proc. of {ICML}}, volume~37, pages 2256--2265, 2015.

\bibitem[Song et~al.(2021{\natexlab{a}})Song, Meng, and
  Ermon]{DBLP:conf/iclr/SongME21}
J.~Song, C.~Meng, and S.~Ermon.
\newblock Denoising diffusion implicit models.
\newblock In \emph{Proc. of {ICLR}}, 2021{\natexlab{a}}.

\bibitem[Song et~al.(2021{\natexlab{b}})Song, Sohl{-}Dickstein, Kingma, Kumar,
  Ermon, and Poole]{DBLP:conf/iclr/0011SKKEP21}
Y.~Song, J.~Sohl{-}Dickstein, D.~P. Kingma, A.~Kumar, S.~Ermon, and B.~Poole.
\newblock Score-based generative modeling through stochastic differential
  equations.
\newblock In \emph{Proc. of {ICLR}}, 2021{\natexlab{b}}.

\bibitem[Song et~al.(2021{\natexlab{c}})Song, Sohl-Dickstein, Kingma, Kumar,
  Ermon, and Poole]{song2021scorebased}
Y.~Song, J.~Sohl-Dickstein, D.~P. Kingma, A.~Kumar, S.~Ermon, and B.~Poole.
\newblock Score-based generative modeling through stochastic differential
  equations.
\newblock \emph{CoRR}, abs/2011.13456, 2021{\natexlab{c}}.

\bibitem[Subramanian et~al.(2022)Subramanian, Annadani, Sheth, Ke, Deleu,
  Bauer, Nowrouzezahrai, and Kahou]{subramanian2022learning}
J.~Subramanian, Y.~Annadani, I.~Sheth, N.~R. Ke, T.~Deleu, S.~Bauer,
  D.~Nowrouzezahrai, and S.~E. Kahou.
\newblock Learning latent structural causal models.
\newblock \emph{CoRR}, abs/2210.13583, 2022.

\bibitem[Traub(2022)]{traub2022representation}
J.~Traub.
\newblock Representation learning with diffusion models.
\newblock \emph{arXiv preprint arXiv:2210.11058}, 2022.

\bibitem[Van~Steenkiste et~al.(2019)Van~Steenkiste, Locatello, Schmidhuber, and
  Bachem]{van2019disentangled}
S.~Van~Steenkiste, F.~Locatello, J.~Schmidhuber, and O.~Bachem.
\newblock Are disentangled representations helpful for abstract visual
  reasoning?
\newblock \emph{Advances in Neural Information Processing Systems}, 32, 2019.

\bibitem[Vincent(2011)]{vincent2011connection}
P.~Vincent.
\newblock A connection between score matching and denoising autoencoders.
\newblock \emph{Neural computation}, 23\penalty0 (7):\penalty0 1661--1674,
  2011.

\bibitem[Von~K{\"u}gelgen et~al.(2021)Von~K{\"u}gelgen, Sharma, Gresele,
  Brendel, Sch{\"o}lkopf, Besserve, and Locatello]{von2021self}
J.~Von~K{\"u}gelgen, Y.~Sharma, L.~Gresele, W.~Brendel, B.~Sch{\"o}lkopf,
  M.~Besserve, and F.~Locatello.
\newblock Self-supervised learning with data augmentations provably isolates
  content from style.
\newblock \emph{Proc. of {N}eur{IPS}}, 34:\penalty0 16451--16467, 2021.

\bibitem[Wang et~al.(2023)Wang, Schiff, Gokaslan, Pan, Wang, De~Sa, and
  Kuleshov]{wang2023infodiffusion}
Y.~Wang, Y.~Schiff, A.~Gokaslan, W.~Pan, F.~Wang, C.~De~Sa, and V.~Kuleshov.
\newblock Infodiffusion: Representation learning using information maximizing
  diffusion models.
\newblock \emph{arXiv preprint arXiv:2306.08757}, 2023.

\bibitem[Wu et~al.(2022)Wu, Dvornik, Greff, Kipf, and Garg]{wu2022slotformer}
Z.~Wu, N.~Dvornik, K.~Greff, T.~Kipf, and A.~Garg.
\newblock Slotformer: Unsupervised visual dynamics simulation with
  object-centric models.
\newblock \emph{CoRR}, abs/2210.05861, 2022.

\bibitem[Yang et~al.(2020)Yang, Liu, Chen, Shen, Hao, and
  Wang]{yang2020causalvae}
M.~Yang, F.~Liu, Z.~Chen, X.~Shen, J.~Hao, and J.~Wang.
\newblock Causalvae: Structured causal disentanglement in variational
  autoencoder.
\newblock \emph{CoRR}, abs/2208.14153, 2020.

\bibitem[Yoon et~al.(2023)Yoon, Wu, Bae, and Ahn]{yoon2023investigation}
J.~Yoon, Y.-F. Wu, H.~Bae, and S.~Ahn.
\newblock An investigation into pre-training object-centric representations for
  reinforcement learning.
\newblock \emph{CoRR}, abs/2302.04419, 2023.

\end{thebibliography}
\newpage
\onecolumn
\renewcommand{\thesection}{\Alph{section}}
\setcounter{section}{0}
\noindent {\LARGE\textbf{Appendix}}
\section{Problem Formulation \& ELBO}
\label{appendix:elbo}

The ELBO for the proposed framework will be (For simplicity, we only derive the ELBO when using single representations independent of time, i.e., $e \in \mathbb{R}^d$. The ELBO for the infinite-dimensional case would be similar):
\begin{align}
    \log p (x, \Tilde{x}) & \geq
    \mathbb{E}_{q(e, \Tilde{e}, u, \Tilde{u}, I | x, \Tilde{x})}
    \left [ 
        \log \frac{p(x, \Tilde{x}, u, \Tilde{u}, e, \Tilde{e}, I)}{q(e, \Tilde{e}, I, u, \Tilde{u}|x, \Tilde{x})} 
    \right ] \nonumber \\
    = & 
    \mathbb{E}_{q(e, \Tilde{e}, u, \Tilde{u}, I | x, \Tilde{x})}
    \left[ 
        \log \frac{p(I)}{q(I|x, \Tilde{x})} + \log \frac{p(e)p(\Tilde{e}|e, I)}
        {q(e, \Tilde{e}|x, \Tilde{x}, I)} + \log \frac{p(x, u|e)}{q(u|x)} + \log \frac{p(\Tilde{x}, \Tilde{u}|\Tilde{e})}{q(\Tilde{u}|\Tilde{x})}
    \right] \nonumber \\
    =&
    \mathbb{E}_{q(I|x, \Tilde{x})} \mathbb{E}_{q(e, \Tilde{e}|x, \Tilde{x}, I)} \mathbb{E}_{q(u|x)} \mathbb{E}_{q(\Tilde{u}|\Tilde{x})}
    \Bigg[
        \Big[
             \log p(I) + \log p(e) + \log p(\Tilde{e}|e, I)
            - \log q(I|x, \Tilde{x}) \nonumber \\ 
            -& \log q(e, \Tilde{e}|x, \Tilde{x}, I)
        \Big]
        +
        \left[
             \log \frac{p(x, u|e)}{q(u|x)} + \log \frac{p(\Tilde{x}, \Tilde{u}|\Tilde{e})}{q(\Tilde{u}|\Tilde{x})}
        \right]
        \nonumber
    \Bigg]
\end{align}
The terms in the first bracket correspond to the intervention encoder and the noise encoding module, respectively, and the terms in the second bracket correspond to the diffusion model conditioned on pre- and post-intervention noise encodings.

\citet{song2021scorebased} shows that the discretization of SDE formulations of the diffusion model is equivalent to discrete-time diffusion models. Therefore, for simplicity, we derive the ELBO for discrete-time diffusion models. Following \citep{luo2022understanding}, for a discrete-time diffusion model where $t \in [1, T]$, we have
\begin{align}
    \mathbb{E}_{q(I|x, \Tilde{x})} & \mathbb{E}_{q(e, \Tilde{e}|x, \Tilde{x}, I)} \mathbb{E}_{q(u|x)} \mathbb{E}_{q(\Tilde{u}|\Tilde{x})}
    \Bigg[
        \log \frac{p(x, u|e)}{q(u|x)}
    \Bigg] \nonumber \\
    & = 
    \mathbb{E}_{q(I|x, \Tilde{x})} \mathbb{E}_{q(e, \Tilde{e}|x, \Tilde{x}, I)} \mathbb{E}_{q(u|x)} \mathbb{E}_{q(\Tilde{u}|\Tilde{x})} 
    \Bigg[ 
        \mathbb{E}_{q(u_1|x)} 
        [\log p(x|u_1)] - D_{KL}(q(u_T|x)||p(u_T)) \nonumber \\
    & - \sum_{t=2}^{T} \mathbb{E}_{q(u_t|x)} [ D_{KL}(q(u_{t-1}|u_t, x, e) || p(u_{t-1}|u_t, e)]
    \Bigg] \label{eq:elbo1}
\end{align}
where we have that
\begin{itemize}[itemsep=4pt,parsep=0pt]
    \item $\mathbb{E}_{q(u_1|x)} [\log p(x|u_1)]$ is the reconstruction term and it can be defined in a way that it is constant so it can be ignored during training;
    \item $D_{KL}(q(u_T|x)||p(u_T))$ is the prior matching term and can similarly be defined in a way that it is constant;
    \item $\mathbb{E}_{u_t|x} [D_{KL}(q(u_{t-1}|u_t, x, e) || p(u_{t-1}|u_t, e)]$ is a denoising matching term. This term is the origin of different interpretations of the score-based diffusion models. 
\end{itemize}    
For the SDE formulation of the forward process, the denoising matching term becomes \citep{song2021scorebased}
    \begin{equation}
    \label{eq:score_matching}
        \lambda(t)||s_\theta(u_t, e, t) - \nabla_{u_t} \log p(u_t|x)||_2^2.
    \end{equation}
The weight $\lambda(t)$ of denoising matching terms is related to the diffusion coefficient of the forward SDE. For a Variance Exploding SDE the weight is defined as $\lambda(t) = 2 \sigma^2(t) \log (\sigma_{max}/\sigma_{min})$ with $\sigma(t) = \sigma_{min} \cdot (\sigma_{max}/\sigma_{min})^t $. 

Therefore, by combining \eqref{eq:elbo1} with \eqref{eq:score_matching}, the ELBO becomes
\begin{align*}
    \log p(x, \Tilde{x}) & \geq
    \mathbb{E}_{p(x,\Tilde{x})} 
        \mathbb{E}_{q(I|x,\Tilde{x})} \mathbb{E}_{q(e, \Tilde{e}|x, \Tilde{x}, I)}
            \mathbb{E}_{t \sim U(0, 1)} \mathbb{E}_{q(u_t|x)}\mathbb{E}_{q(\Tilde{u}_t|\Tilde{x})} \\
            \Biggl[
                &
                    \log p(I) + \log p(e) + \log p(\Tilde{e}|e, I)
                    -
                    \log q(I|x, \Tilde{x})
                    - \log q(e, \Tilde{e}|x, \Tilde{x}, I) 
                \\
                +&
                \lambda(t)
                \Big[
                ||s_\theta(u_t, e, t) - \nabla_{u_t} \log p(u_t|x)||_2^2
                +
                ||s_\theta(\Tilde{u}_t, \Tilde{e}, t) - \nabla_{\Tilde{u}_t} \log p(\Tilde{u}_t|\Tilde{x})||_2^2
                \Big]
            \Biggl]
\end{align*}

For infinite-dimensional representations, we can derive the ELBO using a similar argument. In this case, the formula for the ELBO is
\begin{align*}
    \log p(x, \Tilde{x}) & \geq
    \mathbb{E}_{p(x,\Tilde{x})} 
        \mathbb{E}_{q(I|x,\Tilde{x})} \mathbb{E}_{t \sim U(0, 1)} \mathbb{E}_{q(e_t, \Tilde{e}_t|x, \Tilde{x}, I)} 
            \mathbb{E}_{q(u_t|x)}\mathbb{E}_{q(\Tilde{u}_t|\Tilde{x})} \\
            \Biggl[
            &
                \log p(I) + \log p(e_t)
                +  \log p(\Tilde{e}_t|e_t, I)
                - \log q(I|x, \Tilde{x}) 
                - \log q(e_t, \Tilde{e}_t|x, \Tilde{x}, I) \\ 
                +&
                \lambda(t)||s_\theta(u_t, e_t, t) - \nabla_{u_t} \log p(u_t|x)||_2^2
                +
                \lambda(t)||s_\theta(\Tilde{u}_t, \Tilde{e}_t, t) - \nabla_{\Tilde{u}_t} \log p(\Tilde{u}_t|\Tilde{x})||_2^2
            \Biggl],
\end{align*}

\section{Implementation Details}
\label{appendix:implementation_details}


\paragraph{Training}
For the training, we follow the 4-phase training of \citet{brehmer2022weakly} but consider only the first 3 phases. In summary, we consider the following steps:
\begin{enumerate}[itemsep=4pt,parsep=0pt,label=(\arabic*)]
    \item We begin by training the diffusion model and the encoding module together on data pairs for $20$ epochs. This can be interpreted as a warm-up on the diffusion model and the encoding module to extract meaningful representations of data.
    \item We include all modules except for solution functions. We consider $p(\Tilde{e}_i|e)$ to be a uniform probability density. We do this phase for $50$ epochs.
    \item We include solution functions and train the whole framework with the proposed loss and do this for $50$ epochs.
\end{enumerate}
We find out that considering our data generation process, including the fourth training phase of \citet{brehmer2022weakly} has no impact on the model's performance. Consequently, we choose to disregard it in our analysis. We use the loss in Eq. \ref{eq:loss} as the objective function and consider the coefficient of the regularization term $\mathcal{L}_{entropy}$ to be 1. Therefore, our overall loss function is then given by $\mathcal{L} = \mathcal{L}_{model} + \mathcal{L}_{entropy}$.

\paragraph{Architectures \& Hyperparameters}
We train the model for $120$ epochs and use the learning rate of 3e-4 with a batch size of $64$. $\beta$ is initially set to $0$ and increased to $1$ during training. The noise encoder is considered Gaussian, with mean and standard deviation parameterized as an MLP with two hidden layers and 64 units each and ReLU activation functions. The architecture of the score function of the diffusion model is based on NCSN++ architecture \citep{song2021scorebased} with the same set of hyperparameters. As the input $x$ is 16-dimensional and the score model follows a convolutional architecture, we reshape the input into a $4\times4$ format and then feed it into the diffusion model. Furthermore, In the forward SDE, $\sigma_{min}$ and $\sigma_{max}$ are set to 0.01 and 50, respectively. 

\section{Missing Plots}
\label{appendix:missing_figures}

\begin{figure}[h]
    \centering
         \includegraphics[width=0.92\textwidth]{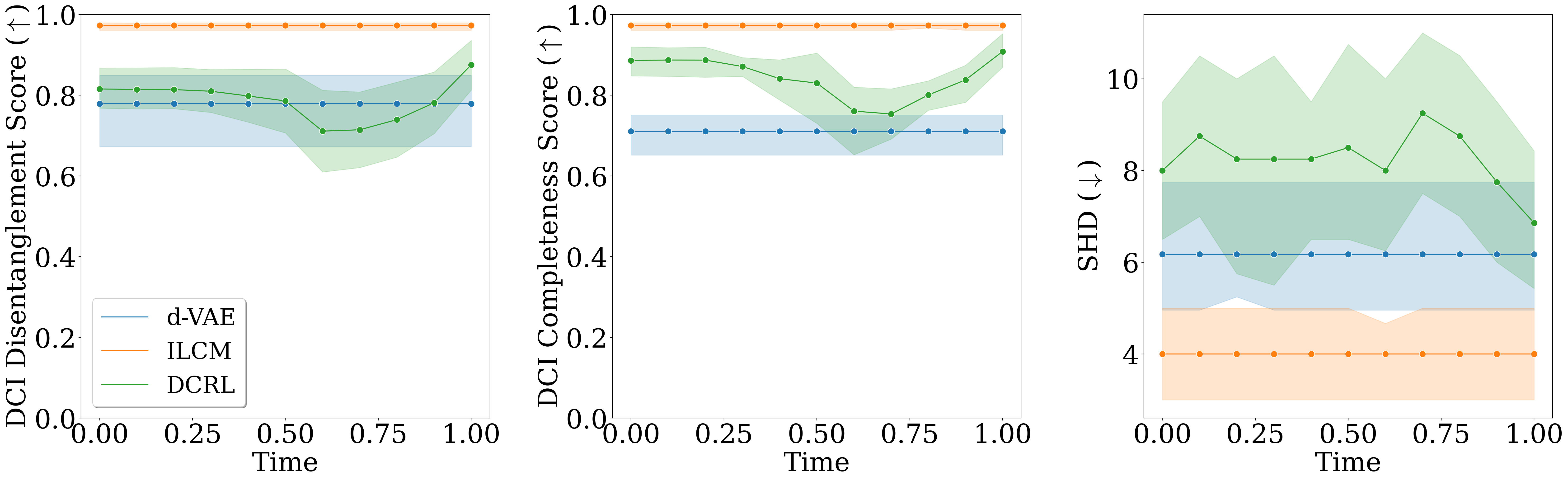}
         \includegraphics[width=0.92\textwidth]{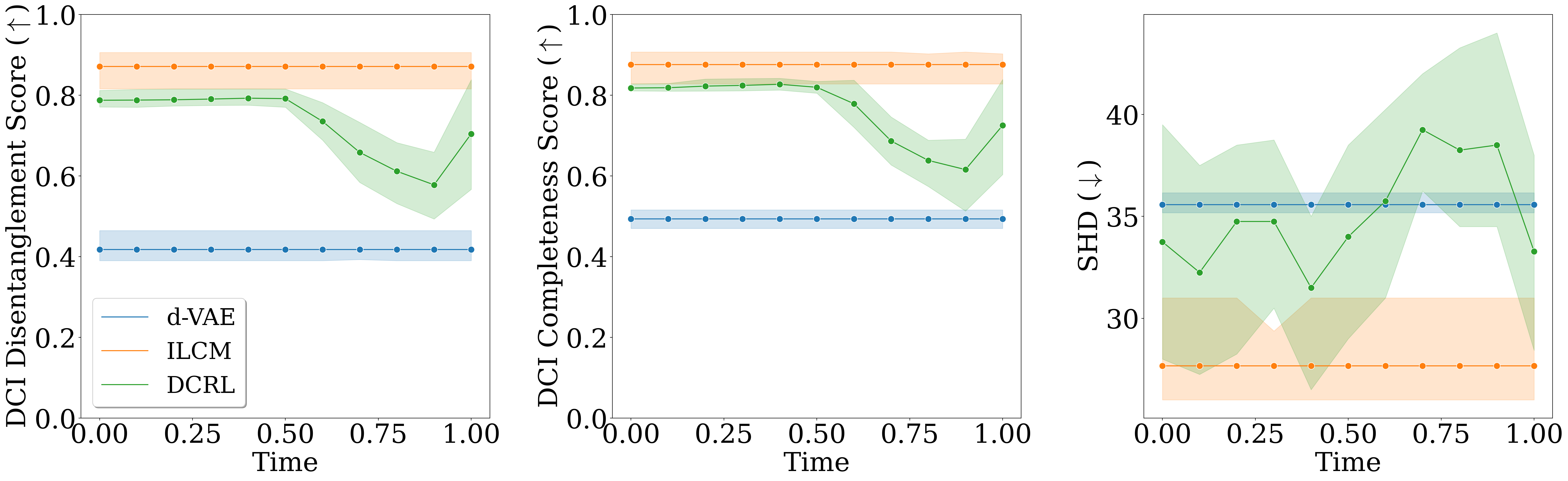}
         \centering
         \includegraphics[width=0.92\textwidth]{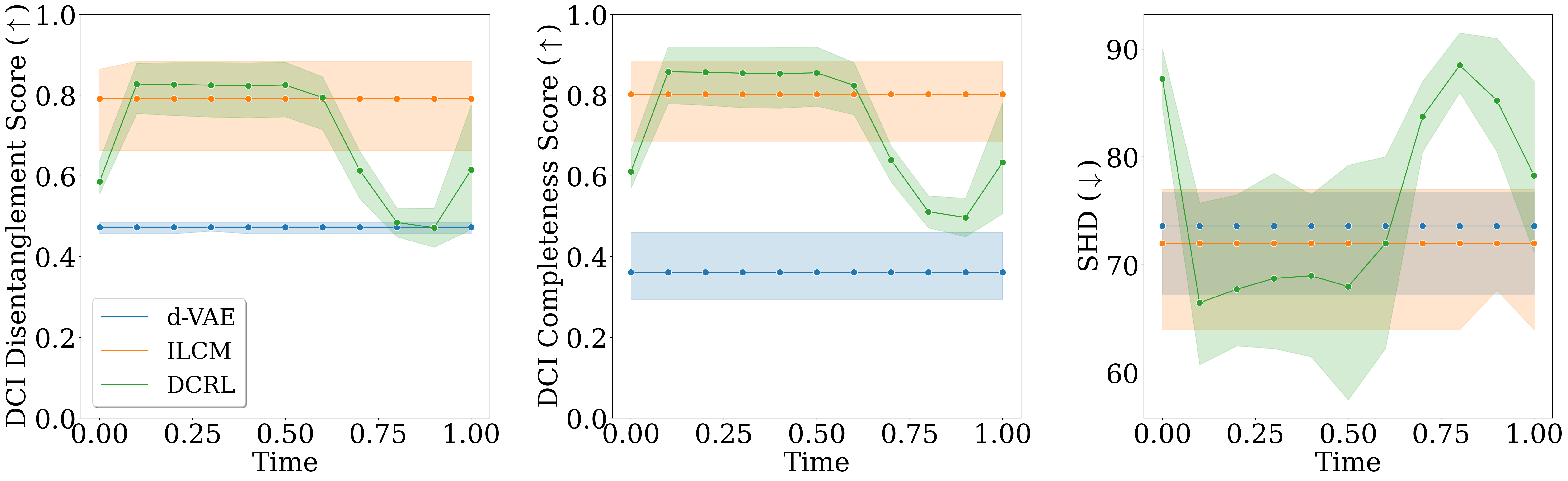}
     \caption{Comparison of models on different metrics when using infinite-dimensional representations. From top to bottom, (a), (b), and (c) correspond to experiments with 5, 10, and 15 causal variables, respectively. We sample points from the trajectory at intervals of $0.1$, creating a total of $11$ specific timesteps. Typically, representations in the middle of the trajectory carry the most information, often matching or surpassing the baseline performance. As we move further in time, representations seem to lose some information, but they improve as they approach the end of the trajectory. Furthermore, the framework performs worse or on par with baselines in lower dimensions but generally outperforms them in higher dimensions.}
     \label{fig:metrics_infinite_dimensional}
\end{figure}
\end{document}